\documentclass[10pt,twocolumn,letterpaper]{article}
\makeatletter
\renewcommand\paragraph{\@startsection{paragraph}{4}{\z@}{1ex}{-1em}{\normalfont\normalsize\bfseries}}
\makeatother
\usepackage{iccv}
\usepackage{times}
\usepackage{epsfig}
\usepackage{graphicx}
\usepackage{amsmath}
\usepackage{colortbl}
\usepackage{amssymb}
\usepackage{adjustbox}

\usepackage{times}
\usepackage{epsfig}
\usepackage{graphicx}
\usepackage{amsmath}
\usepackage{amssymb}
\usepackage{multirow}

\usepackage[utf8]{inputenc} 
\usepackage[T1]{fontenc}    
\usepackage{url}            
\usepackage{booktabs}       
\usepackage{amsfonts}       
\usepackage{nicefrac}       
\usepackage{bbm}            
\usepackage{enumitem}
\usepackage{float}
\usepackage{microtype}
\usepackage[ruled,vlined,linesnumbered]{algorithm2e}
\makeatletter
\@namedef{ver@everyshi.sty}{}
\makeatother

\usepackage[dvipsnames,svgnames,x11names]{xcolor}
\usepackage{tikz}
\usetikzlibrary{arrows.meta,shapes,calc,matrix,fit,positioning,backgrounds,decorations.markings}
\usepackage{pgfplots}
\usepackage{pgfplotstable}
\pgfplotsset{compat=1.9}
\usepackage{xstring}

\usepgfplotslibrary{external}
\newcommand{\extfig}[2]{\tikzsetnextfilename{#1}{#2}}

\IfBeginWith*{\jobname}{fig/extern/}{\finalcopy}{}

\tikzstyle{every picture}+=[
	remember picture,
	every text node part/.style={align=center},
	every matrix/.append style={ampersand replacement=\&},
]
\tikzstyle{tight} = [inner sep=0pt,outer sep=0pt]
\tikzstyle{node}  = [draw,circle,tight,minimum size=12pt,anchor=center]
\tikzstyle{op}    = [draw,circle,tight]
\tikzstyle{dot}   = [fill,draw,circle,inner sep=1pt,outer sep=0]
\tikzstyle{pt}    = [fill,draw,circle,inner sep=1.5pt,outer sep=.2pt]
\tikzstyle{box}   = [draw,thick,rectangle,inner sep=3pt]
\tikzstyle{high}  = [black!60]
\tikzstyle{group} = [high,box,opacity=.5]
\tikzstyle{rectc} = [tight,transform shape]
\tikzstyle{rect}  = [rectc,anchor=south west]

\newcommand{\leg}[1]{\addlegendentry{#1}}

\tikzset{every mark/.append style={solid}}
\pgfplotsset{
	grid=both, width=\columnwidth, try min ticks=5,
	every axis/.append style={font=\small},
	every axis plot/.append style={thick,mark=none,mark size=1.8,tension=0.18},
	legend cell align=left, legend style={fill opacity=0.8},
	xticklabel={\pgfmathprintnumber[assume math mode=true]{\tick}},
	yticklabel={\pgfmathprintnumber[assume math mode=true]{\tick}},
	nodes near coords math/.style={
		nodes near coords={\pgfmathprintnumber[assume math mode=true]{\pgfplotspointmeta}},
	},
}

\pgfplotsset{
	dash/.style={mark=o,dashed,opacity=0.6},
	dott/.style={mark=o,dotted,opacity=0.6},
	nolim/.style={enlargelimits=false},
	plain/.style={every axis plot/.append style={},nolim,grid=none},
}

\usepackage[pagebackref=true,breaklinks=true,letterpaper=true,colorlinks,bookmarks=false]{hyperref}

\iccvfinalcopy 

\def\iccvPaperID{****} 

\ificcvfinal\fi

\begin{document}

\title{Adaptive manifold for imbalanced transductive few-shot learning}


\author{Michalis Lazarou$^1$\ \ \ \ Yannis Avrithis$^2$\ \  \  \ Tania Stathaki$^1$\\
Imperial College London$^1$\\
Institute of Advanced Research on Artificial Intelligence$^2$
}

\maketitle
\ificcvfinal\thispagestyle{empty}\fi

\newcommand{\head}[1]{{\smallskip\noindent\textbf{#1}}}
\newcommand{\alert}[1]{{\color{red}{#1}}}
\newcommand{\sm}{\scriptsize}
\newcommand{\eq}[1]{(\ref{eq:#1})}

\newcommand{\Th}[1]{\textsc{#1}}
\newcommand{\mr}[2]{\multirow{#1}{*}{#2}}
\newcommand{\mc}[2]{\multicolumn{#1}{c}{#2}}
\newcommand{\tb}[1]{\textbf{#1}}
\newcommand{\ch}{\checkmark}

\newcommand{\red}[1]{{\color{red}{#1}}}
\newcommand{\blue}[1]{{\color{blue}{#1}}}
\newcommand{\green}[1]{{\color{green}{#1}}}
\newcommand{\gray}[1]{{\color{gray}{#1}}}

\newcommand{\citeme}[1]{\red{[XX]}}
\newcommand{\refme}[1]{\red{(XX)}}

\newcommand{\fig}[2][1]{\includegraphics[width=#1\columnwidth]{fig/#2}}
\newcommand{\figh}[2][1]{\includegraphics[height=#1\columnwidth]{fig/#2}}


\newcommand{\tran}{^\top}
\newcommand{\mtran}{^{-\top}}
\newcommand{\zcol}{\mathbf{0}}
\newcommand{\zrow}{\zcol\tran}

\newcommand{\ind}{\mathbbm{1}}
\newcommand{\expect}{\mathbb{E}}
\newcommand{\nat}{\mathbb{N}}
\newcommand{\zahl}{\mathbb{Z}}
\newcommand{\real}{\mathbb{R}}
\newcommand{\proj}{\mathbb{P}}
\newcommand{\prob}{\mathbf{Pr}}
\newcommand{\normal}{\mathcal{N}}

\newcommand{\mif}{\textrm{if}\ }
\newcommand{\other}{\textrm{otherwise}}
\newcommand{\minimize}{\textrm{minimize}\ }
\newcommand{\maximize}{\textrm{maximize}\ }
\newcommand{\st}{\textrm{subject\ to}\ }

\newcommand{\id}{\operatorname{id}}
\newcommand{\const}{\operatorname{const}}
\newcommand{\sgn}{\operatorname{sgn}}
\newcommand{\var}{\operatorname{Var}}
\newcommand{\mean}{\operatorname{mean}}
\newcommand{\trace}{\operatorname{tr}}
\newcommand{\diag}{\operatorname{diag}}
\newcommand{\vect}{\operatorname{vec}}
\newcommand{\cov}{\operatorname{cov}}
\newcommand{\sign}{\operatorname{sign}}
\newcommand{\prj}{\operatorname{proj}}

\newcommand{\softmax}{\operatorname{softmax}}
\newcommand{\clip}{\operatorname{clip}}

\newcommand{\defn}{\mathrel{:=}}
\newcommand{\peq}{\mathrel{+\!=}}
\newcommand{\meq}{\mathrel{-\!=}}

\newcommand{\floor}[1]{\left\lfloor{#1}\right\rfloor}
\newcommand{\ceil}[1]{\left\lceil{#1}\right\rceil}
\newcommand{\inner}[1]{\left\langle{#1}\right\rangle}
\newcommand{\norm}[1]{\left\|{#1}\right\|}
\newcommand{\abs}[1]{\left|{#1}\right|}
\newcommand{\frob}[1]{\norm{#1}_F}
\newcommand{\card}[1]{\left|{#1}\right|\xspace}
\newcommand{\diff}{\mathrm{d}}
\newcommand{\der}[3][]{\frac{d^{#1}#2}{d#3^{#1}}}
\newcommand{\pder}[3][]{\frac{\partial^{#1}{#2}}{\partial{#3^{#1}}}}
\newcommand{\ipder}[3][]{\partial^{#1}{#2}/\partial{#3^{#1}}}
\newcommand{\dder}[3]{\frac{\partial^2{#1}}{\partial{#2}\partial{#3}}}

\newcommand{\wb}[1]{\overline{#1}}
\newcommand{\wt}[1]{\widetilde{#1}}

\def\xssp{\hspace{1pt}}
\def\ssp{\hspace{3pt}}
\def\msp{\hspace{5pt}}
\def\lsp{\hspace{12pt}}

\newcommand{\cA}{\mathcal{A}}
\newcommand{\cB}{\mathcal{B}}
\newcommand{\cC}{\mathcal{C}}
\newcommand{\cD}{\mathcal{D}}
\newcommand{\cE}{\mathcal{E}}
\newcommand{\cF}{\mathcal{F}}
\newcommand{\cG}{\mathcal{G}}
\newcommand{\cH}{\mathcal{H}}
\newcommand{\cI}{\mathcal{I}}
\newcommand{\cJ}{\mathcal{J}}
\newcommand{\cK}{\mathcal{K}}
\newcommand{\cL}{\mathcal{L}}
\newcommand{\cM}{\mathcal{M}}
\newcommand{\cN}{\mathcal{N}}
\newcommand{\cO}{\mathcal{O}}
\newcommand{\cP}{\mathcal{P}}
\newcommand{\cQ}{\mathcal{Q}}
\newcommand{\cR}{\mathcal{R}}
\newcommand{\cS}{\mathcal{S}}
\newcommand{\cT}{\mathcal{T}}
\newcommand{\cU}{\mathcal{U}}
\newcommand{\cV}{\mathcal{V}}
\newcommand{\cW}{\mathcal{W}}
\newcommand{\cX}{\mathcal{X}}
\newcommand{\cY}{\mathcal{Y}}
\newcommand{\cZ}{\mathcal{Z}}
\newcommand{\cPi}{\mathcal{\pi}}

\newcommand{\vA}{\mathbf{A}}
\newcommand{\vB}{\mathbf{B}}
\newcommand{\vC}{\mathbf{C}}
\newcommand{\vD}{\mathbf{D}}
\newcommand{\vE}{\mathbf{E}}
\newcommand{\vF}{\mathbf{F}}
\newcommand{\vG}{\mathbf{G}}
\newcommand{\vH}{\mathbf{H}}
\newcommand{\vI}{\mathbf{I}}
\newcommand{\vJ}{\mathbf{J}}
\newcommand{\vK}{\mathbf{K}}
\newcommand{\vL}{\mathbf{L}}
\newcommand{\vM}{\mathbf{M}}
\newcommand{\vN}{\mathbf{N}}
\newcommand{\vO}{\mathbf{O}}
\newcommand{\vP}{\mathbf{P}}
\newcommand{\vQ}{\mathbf{Q}}
\newcommand{\vR}{\mathbf{R}}
\newcommand{\vS}{\mathbf{S}}
\newcommand{\vT}{\mathbf{T}}
\newcommand{\vU}{\mathbf{U}}
\newcommand{\vV}{\mathbf{V}}
\newcommand{\vW}{\mathbf{W}}
\newcommand{\vX}{\mathbf{X}}
\newcommand{\vY}{\mathbf{Y}}
\newcommand{\vZ}{\mathbf{Z}}

\newcommand{\va}{\mathbf{a}}
\newcommand{\vb}{\mathbf{b}}
\newcommand{\vc}{\mathbf{c}}
\newcommand{\vd}{\mathbf{d}}
\newcommand{\ve}{\mathbf{e}}
\newcommand{\vf}{\mathbf{f}}
\newcommand{\vg}{\mathbf{g}}
\newcommand{\vh}{\mathbf{h}}
\newcommand{\vi}{\mathbf{i}}
\newcommand{\vj}{\mathbf{j}}
\newcommand{\vk}{\mathbf{k}}
\newcommand{\vl}{\mathbf{l}}
\newcommand{\vm}{\mathbf{m}}
\newcommand{\vn}{\mathbf{n}}
\newcommand{\vo}{\mathbf{o}}
\newcommand{\vp}{\mathbf{p}}
\newcommand{\vq}{\mathbf{q}}
\newcommand{\vr}{\mathbf{r}}
\newcommand{\Vs}{\mathbf{s}}
\newcommand{\vt}{\mathbf{t}}
\newcommand{\vu}{\mathbf{u}}
\newcommand{\vv}{\mathbf{v}}
\newcommand{\uu}{\mathbf{u}}
\newcommand{\cc}{\mathbf{c}}
\newcommand{\vw}{\mathbf{w}}
\newcommand{\vx}{\mathbf{x}}
\newcommand{\vy}{\mathbf{y}}
\newcommand{\vz}{\mathbf{z}}

\newcommand{\vone}{\mathbf{1}}
\newcommand{\vzero}{\mathbf{0}}

\newcommand{\valpha}{{\boldsymbol{\alpha}}}
\newcommand{\vbeta}{{\boldsymbol{\beta}}}
\newcommand{\vgamma}{{\boldsymbol{\gamma}}}
\newcommand{\vdelta}{{\boldsymbol{\delta}}}
\newcommand{\vepsilon}{{\boldsymbol{\epsilon}}}
\newcommand{\vzeta}{{\boldsymbol{\zeta}}}
\newcommand{\veta}{{\boldsymbol{\eta}}}
\newcommand{\vtheta}{{\boldsymbol{\theta}}}
\newcommand{\viota}{{\boldsymbol{\iota}}}
\newcommand{\vkappa}{{\boldsymbol{\kappa}}}
\newcommand{\vlambda}{{\boldsymbol{\lambda}}}
\newcommand{\vmu}{{\boldsymbol{\mu}}}
\newcommand{\vnu}{{\boldsymbol{\nu}}}
\newcommand{\vxi}{{\boldsymbol{\xi}}}
\newcommand{\vomikron}{{\boldsymbol{\omikron}}}
\newcommand{\vpi}{{\boldsymbol{\pi}}}
\newcommand{\vrho}{{\boldsymbol{\rho}}}
\newcommand{\vsigma}{{\boldsymbol{\sigma}}}
\newcommand{\vtau}{{\boldsymbol{\tau}}}
\newcommand{\vupsilon}{{\boldsymbol{\upsilon}}}
\newcommand{\vphi}{{\boldsymbol{\phi}}}
\newcommand{\vchi}{{\boldsymbol{\chi}}}
\newcommand{\vpsi}{{\boldsymbol{\psi}}}
\newcommand{\vomega}{{\boldsymbol{\omega}}}

\newcommand{\rLambda}{\mathrm{\Lambda}}
\newcommand{\rSigma}{\mathrm{\Sigma}}

\newcommand{\vLambda}{\bm{\rLambda}}
\newcommand{\vSigma}{\bm{\rSigma}}

\makeatletter
\newcommand*\bdot{\mathpalette\bdot@{.7}}
\newcommand*\bdot@[2]{\mathbin{\vcenter{\hbox{\scalebox{#2}{$\m@th#1\bullet$}}}}}
\makeatother

\makeatletter
\DeclareRobustCommand\onedot{\futurelet\@let@token\@onedot}
\def\@onedot{\ifx\@let@token.\else.\null\fi\xspace}

\def\eg{\emph{e.g}\onedot} \def\Eg{\emph{E.g}\onedot}
\def\ie{\emph{i.e}\onedot} \def\Ie{\emph{I.e}\onedot}
\def\cf{\emph{cf}\onedot} \def\Cf{\emph{Cf}\onedot}
\def\etc{\emph{etc}\onedot} \def\vs{\emph{vs}\onedot}
\def\wrt{w.r.t\onedot} \def\dof{d.o.f\onedot} \def\aka{a.k.a\onedot}
\def\etal{\emph{et al}\onedot}
\makeatother

\newcommand{\base}{\mathrm{base}}
\newcommand{\novel}{\mathrm{novel}}
\newcommand{\NN}{\mathrm{NN}}
\newcommand{\masked}{\mathrm{masked}}
\newcommand{\soft}{\mathrm{soft}}
\newcommand{\ce}{\mathrm{CE}}
\newcommand{\bal}{\mathrm{bal}}
\newcommand{\imbal}{\mathrm{imbal}}

\newcommand{\Dir}{\operatorname{Dir}}

\newcommand{\ours}{AM\xspace}
\newcommand{\oursplc}{AM$_{\textsc{plc}}$\xspace}
\newcommand{\plc}{PLC\xspace}
\newcommand{\ilpc}{iLPC\xspace}

\begin{abstract}
Transductive few-shot learning algorithms have showed substantially superior performance over their inductive counterparts by leveraging the unlabeled queries. However, the vast majority of such methods are evaluated on perfectly class-balanced benchmarks. It has been shown that they undergo remarkable drop in performance under a more realistic, imbalanced setting.

To this end, we propose a novel algorithm to address imbalanced transductive few-shot learning, named \emph{Adaptive Manifold}. Our method exploits the underlying manifold of the labeled support examples and unlabeled queries by using manifold similarity to predict the class probability distribution per query. It is parameterized by one centroid per class as well as a set of graph-specific parameters that determine the manifold. All parameters are optimized through a loss function that can be tuned towards class-balanced or imbalanced distributions. The manifold similarity shows substantial improvement over Euclidean distance, especially in the 1-shot setting.

Our algorithm outperforms or is on par with other state of the art methods in three benchmark datasets, namely \emph{mini}ImageNet, \emph{tiered}ImageNet and CUB, and three different backbones, namely ResNet-18, WideResNet-28-10 and DenseNet-121. In certain cases, our algorithm outperforms the previous state of the art by as much as $4.2\%$.
\end{abstract}

\section{Introduction}
\label{sec:intro}

\begin{figure*}[h]
\centering
\includegraphics[width=\textwidth,height=.22\textwidth]{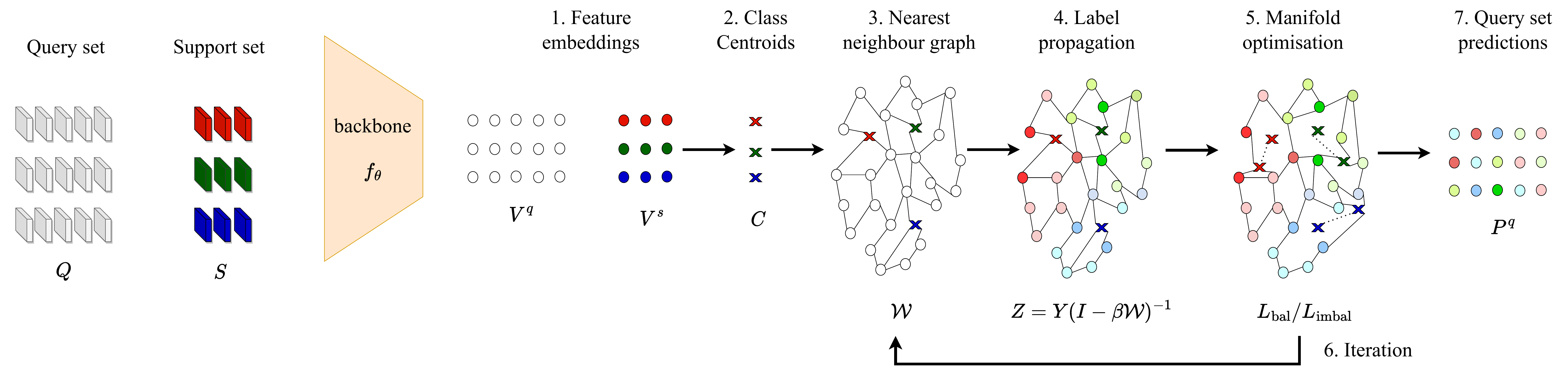}
\vspace{0pt}
\caption{\emph{Overview of our method}. 1) Given a support set, $S$, and a query set, $Q$, we extract features $V^s$ and $V^q$ using the pre-trained backbone $f_{\theta}$. 2) We calculate class centroids, $C$, using \eq{centroid}. 3) We calculate the $k$-nearest neighbour graph using \eq{edges}, \eq{affinity}, \eq{adj_B} and \eq{adj}. 4) We perform label propagation using \eq{lp}. 5) We optimize the manifold parameters, $\Phi$, using \eq{balanced} or \eq{imbalanced}. 6) We iterate the procedure from graph construction for $r$ steps. 7) We predict pseudo-labels using $P^q$.}
\label{fig:overview}
\end{figure*}

One of the fundamental challenges of deep learning is its reliance on large labeled datasets. Even though weak or self-supervision is gaining momentum, an even greater challenge is the difficulty of obtaining the data itself, even unlabeled. This is the case in applications where the data is scarce, for example in rare animal species~\cite{bertinetto}.

The \emph{few-shot learning} paradigm has attracted significant interest because it investigates the question of how to make deep learning models acquire knowledge from limited data~\cite{matchingNets, prototypical, MAML}. Different methodologies have been proposed to address few-shot learning such as \emph{meta-learning}~\cite{prototypical, MAML, siamese}, \emph{transfer learning}~\cite{rfs, manifoldmixup, denseclassification} and \emph{synthetic data generation}~\cite{AFHN, VIFSL, TFH}. The vast majority of these methods focus on the \emph{inductive} setting, where the assumption is that at inference, every query example is classified independently of the others.

Recent studies have explored the \emph{transductive} few-shot learning setting, where all query examples can be exploited together at test time, showing remarkable improvement in performance~\cite{ilpc, pt_map, lrici, ease, oblique_manifold}.
Some approaches exploit all query examples at the same time by utilizing the data manifold through label propagation~\cite{ilpc} or through the properties of the oblique manifold~\cite{oblique_manifold}. Other approaches utilize the available query examples to improve the class centroids by specialized loss functions \cite{TIM}, using soft K-means \cite{pt_map} or by minimizing the cross-class and intra-class variance\cite{BDCSPN}.

While the query set of transductive few-shot learning benchmarks is unlabeled, it is still curated in the sense that the tasks are perfectly class-balanced. Several state of the art methods are in fact based on this assumption and use class balancing approaches to improve their performance~\cite{ilpc, ease, pt_map, TIM}. However, it has been argued that this is not a realistic setting~\cite{alphaTIM}. As a way to address this flaw, the latter study introduced a new \emph{imbalanced transductive few-shot learning} setting, comparing numerous state of the art methods under a fair setting and showing that their performance drops dramatically.


In this work, focusing on this imbalanced transductive setting~\cite{alphaTIM}, we introduce a new algorithm, called \emph{Adaptive Manifold} (AM), that combines the merits of class centroid approaches and data manifold exploitation approaches. In particular, as illustrated in \autoref{fig:overview}, we initialize the class centroids from the labeled support examples and we propagate the labels along the data manifold, using a $k$-nearest neighbour graph~\cite{semilpavrithis}. Using the loss function proposed in \cite{alphaTIM} we iteratively update both the class centroids as well as the graph parameters. Our algorithm outperforms other state of the art methods in the imbalanced transductive few-shot learning setting.

In summary, we make the following contributions:
\begin{itemize}[itemsep=2pt, parsep=0pt, topsep=3pt]
    \item We are the first, to the best of our knowledge, to obtain class centroids through manifold class similarity on a $k$-nearest neighbour graph and optimize jointly the class centroids along with graph-specific parameters.
    \item We achieve new state of the art performance on the imbalanced transductive few-shot setting under multiple benchmark datasets and networks, outperforming by as much as $4.2\%$ the previous state of the art in the 1-shot setting.
    \item Our method can also perform on par or even outperform many state of the art methods in the standard balanced transductive few-shot setting.
\end{itemize}
\section{Related work}
\label{sec:related}

\subsection{Few-shot learning}

Learning from limited data is a long-standing problem~\cite{one_shot_start}. A large number of the current methods focus on the \emph{meta-learning} paradigm. These can be grouped into three directions: model-based~\cite{SNAIL, santoro, metaNet, hypernetworks}, optimization based~\cite{MAML, reptile, iMAML, LEO} and metric-based~\cite{matchingNets, prototypical, siamese, relationnet}. \emph{Model-based} methods utilize specialized networks such as memory augmented~\cite{santoro} and meta-networks \cite{metaNet} to aid the meta-learning process. \emph{Optimization-based} methods focus on learning a robust model initialization, through gradient-based solutions~\cite{MAML, reptile}, closed-form solutions~\cite{bertinetto} or an LSTM~\cite{ravilstm}. \emph{Metric-based} approaches operate in the embedding space, based on similarities of a query example with class centroids~\cite{prototypical}, using a learned similarity function~\cite{relationnet} or a Siamese network to compare image pairs~\cite{siamese}.

Recent works~\cite{rfs, closerlook} have shown that the transfer learning paradigm can outperform meta-learning methods. \emph{Transfer learning} methods decouple the training from the inference stage and aim at learning powerful representations through the use of well-designed pre-training regimes to train the backbone network. This often involves auxiliary loss functions along with the standard cross entropy loss, such as such as \emph{knowledge-distillation}~\cite{rfs}, mixup-based \emph{data augmentation}~\cite{manifoldmixup} and \emph{self-supervision}, such as predicting rotations~\cite{gidarisRot} and contrastive learning~\cite{fsl_contrastive}.

Another way to address the data deficiency is to augment the support set with synthetic data. \emph{Synthetic data} can be generated either in the image space or in the feature space, by using a \emph{hallucinator} trained on the base classes. The hallucinator can be trained using common generative models, such as \emph{generative adversarial networks} (GANs)~\cite{metaGAN, AFHN, funit} and \emph{variational autoencoders} (VAEs)~\cite{VIFSL}. Hallucinators have also been specifically designed for the few-shot learning paradigm~\cite{Ideme, salnet, DTN, TFH}.

\subsection{Transductive few-shot learning}

Transductive few-shot learning studies the case where all queries are available at inference time and can be exploited to improve predictions. Several methods exploit the data manifold, by using label propagation~\cite{ilpc, local_lp} or embedding propagation~\cite{embeddingpropagation}, or are based on Riemannian geometry, by using the oblique manifold~\cite{oblique_manifold}. Another direction is to use both labeled and unlabelled examples to improve the class centroids. For example, one may use soft $k$-means to iteratively update the class centroids~\cite{pt_map}, rectify prototypes by minimizing the inter-class and intra-class variance~\cite{BDCSPN}, or iteratively adapt class centroids by maximizing the mutual information between query features and their label predictions~\cite{TIM}. It is has also been proposed to iteratively select the most confident pseudo-labeled queries, for example by interpreting this problem as label denoising~\cite{ilpc} or by calculating the credibility of each pseudo-label~\cite{lrici}.

\subsection{Class balancing}
\label{sec:cls_bal}

The commonly used transductive few-shot learning benchmarks use perfectly class-balanced tasks~\cite{TIM}. Several methods exploit this bias by encouraging class-balanced predictions over queries, thereby improving their performance. One way is to optimize, through the Sinkhorn-Knopp algorithm, the query probability matrix, $P$, to have specific row and column sums $\vp$ and $\vq$ respectively~\cite{pt_map,ilpc}. The row sum $\vp$ amounts to the probability distribution of every query, while the column sum corresponds to the total number of queries per class. Another way is to maximize the entropy of the marginal distribution of predicted labels over queries, thus encouraging it to follow a uniform distribution~\cite{TIM}.

However, the authors of \cite{alphaTIM} argue that using perfectly class-balanced tasks is unrealistic. They propose a more realistic imbalanced setting and protocol, benchmarking the performance of several methods. They also introduce a relaxed version of \cite{TIM} based on $\alpha$-divergence, which can effectively address class-imbalanced tasks.

\section{Method}
\label{sec:method}

\subsection{Problem formulation}
\label{sec:problem}

\paragraph{Representation learning}
We assume access to a \emph{base dataset} $D_{\base} = \{(x_i, \vy_i)\}_{i=1}^B$ of $B$ images, where each image $x_i$ has a corresponding one-hot encoded label $\vy_i$ over a set of \emph{base classes} $C_{\base}$. Denoting by $\cX$ the image space, we assume access to a network $f_{\theta}: \cX \to \real^d$ has been trained on $D_{\base}$, which maps an image $x \in \cX$ to an embedding $f_{\theta}(x) \in \real^d$. 

\paragraph{Inference}
We assume access to a \emph{novel dataset} $D_{\novel}$ consisting of images with corresponding labels from a set $C_{\novel}$ of \emph{novel classes}, where $C_{\novel} \cap C_{\base} = \emptyset$. We sample $N$-way $K$-shot tasks, each consisting of a labeled support set, $S = \{(x_i^s, \vy_i^s)\}_{i=1}^L$, where each image $x_i^s$ has a corresponding one-hot encoded label $\vy_i^s = (y_{ji}^s)_{j=1}^N \in \{0,1\}^N$ over $C_{\novel}$, with $N$ novel classes in total and $K$ examples per class, such that the number of examples in $S$ is $L = \card{S} = NK$. We focus on the transductive setting, therefore a task also contains an unlabeled \emph{query set} $Q = \{x_i^q\}_{i=1}^M$ sampled from the same $N$ classes as the support set $S$ where the number of examples in $Q$ is $M = |Q|$. 

\paragraph{Feature extraction}
Given a novel task, we embed all images in $S$ and $Q$ using $f_{\theta}$ and a feature pre-processing function $\eta: \real^d \to \real^d$, to be discussed in \autoref{sec:experiments}. Let $V^s = (\vv_1^s\ \cdots\ \vv_L^s)$ be the $d \times L$ matrix containing the embeddings of $S$, where $\vv_i^s = \eta(f_\theta(x_i^s)) \in \real^d$. Similarly, let $V^q = (\vv_1^q\ \cdots\ \vv_M^q)$ be the $d \times M$ matrix containing the embeddings of $Q$, where $\vv_i^q = \eta(f_\theta(x_i^q)) \in \real^d$.  We also represent $V^s, V^q$ as sets $\cV^s = \{\vv_i^s\}_{i=1}^L$, $\cV^q = \{\vv_i^q\}_{i=1}^M$. Both sets remain fixed in our method.


\subsection{Class centroids}
\label{sec:centroids}

Following~\cite{prototypical}, we define a class centroid $\vc_j \in \real^d$ in the embedding space for each class $j$ in the support set $S$. The centroids are learnable variables but initialized by standard class prototypes~\cite{prototypical}. That is, the centroid $\vc_j$ of class $j$ is initialized by the mean
\begin{equation}
	\vv_j^c = \frac{1}{K} \sum_{\vv_i^s \in \cV^s} y_{ji}^s \vv_i^s
\label{eq:centroid}
\end{equation}
of support embeddings assigned to class $j$. Let $C = (\vc_1\ \cdots\ \vc_N)$ be the $d \times N$ matrix containing the learnable centroids of all $N$ support classes. We also represent $C$ as a set $\cC = \{\vc_j\}_{j=1}^N$.


\subsection{Nearest neighbour graph}
\label{sec:graph}

We collect centroids, support and query embeddings in a single $d \times T$ matrix
\begin{equation}
	V = (\vv_1\ \cdots\ \vv_T) = (C\ \ V^s\ \ V^q),
\label{eq:vertices}
\end{equation}
where $T = N + L + M$. We also represent $V$ as a set $\cV = \{\vv_i\}_{i=1}^T$. Following \cite{semilpavrithis, ilpc}, we construct a $k$-nearest neighbour graph of $\cV$. We define edges between distinct nearest neighbours in $\cV$ that are not both centroids:
\begin{equation}
	E = \{ (\vv_i, \vv_j) \in \cV^2 \setminus \cC^2: \vv_i \in \NN_k(\vv_j)\},
\label{eq:edges}
\end{equation}
where $\NN_k(\vv)$ is the set of $k$-nearest neighbours of $\vv$ in $\cV$, excluding $\vv$. Given $E$, we define the $T \times T$ \emph{affinity matrix} $A = (a_{ij})$ as
\begin{equation}
	a_{ij} =
	\begin{cases}
		\exp\left(-\frac{\norm{\vv_i-\vv_j}^2}{g_{ij} \sigma^2}\right),
			& \mif (\vv_i, \vv_j) \in E \\
		0, & \other,
	\end{cases}
\label{eq:affinity}
\end{equation}
where $g_{ij}$ is a learnable pairwise scaling factor for every pair $(\vv_i, \vv_j)$, collectively represented by $T \times T$ matrix $G = (g_{ij})$, and $\sigma^2$  is a global scaling factor set equal to the standard deviation of $\norm{\vv_i-\vv_j}^2$ for $(\vv_i, \vv_j) \in \cV^2$ as in \cite{embeddingpropagation}. We symmetrize $A$ into the $T \times T$ \emph{adjacency matrix}, $W = \frac{1}{2} (A + A\tran)$. We calculate $W_B$ which is a scaled version of $W$ defined as:
\begin{equation}
	W_B = W \circ B
\label{eq:adj_B}
\end{equation}
where $B \in [0,1)^{T \times T}$ is a learnable $T \times T$ matrix and $\circ$ is the Hadamard product. We normalize $W_B$ by
\begin{equation}
	\cW = D^{-1/2} W_B D^{-1/2},
\label{eq:adj}
\end{equation}
where $D = \diag(W_B \vone_T)$ is the $T \times T$ degree matrix of $W_B$.


\subsection{Label Propagation}
\label{sec:lp}

\paragraph{Labels}

Following~\cite{ZBL+03}, we define the $N \times T$ \emph{label matrix}
\begin{equation}
	Y = (Y^c\ \ Y^s\ \ Y^q) = (I_N\ \ \vzero_{N \times L}\ \ \vzero_{N \times M}).
\label{eq:label}
\end{equation}
That is, $Y$ has one row per class and one column per example, which is an one-hot encoded label for every class centroid in $\cC$ and a zero vector for both support embeddings $\cV^s$ and query embeddings $\cV^q$.

\paragraph{Label propagation}

Given the graph represented by $\cW$ and the label matrix $Y$, label propagation amounts to
\begin{equation}
	Z = Y (I - \beta \cW)^{-1},
\label{eq:lp}
\end{equation}
where $\beta \in [0,1)$ is a scalar hyperparameter that is referred to as $\alpha$ in the standard label propagation~\cite{ZBL+03}.

\paragraph{Predicted probabilities}

The resulting $N \times T$ matrix $Z = (\vz_1 \cdots \vz_T)$ is called \emph{manifold class similarity} matrix, in the sense that column $\vz_i \in \real^N$ expresses how similar embedding vector $\vv_i$ is to each of the $N$ support classes. By taking softmax over columns
\begin{equation}
	\vp_i = \frac{\exp(\tau \vz_i)}{\sum_{j=1}^N \exp(\tau z_{ji})},
\label{eq:softmax}
\end{equation}
with $\tau > 0$ being a positive scale hyperparameter, we define the $N \times T$ \emph{probability matrix}
\begin{equation}
	P = (\vp_1\ \cdots\ \vp_T) = (P^c\ \ P^s\ \ P^q).
\label{eq:prob}
\end{equation}
Matrix $P$ expresses the predicted probability distributions over the support classes. If $P = (p_{ji})$, element $p_{ji}$ expresses the predicted probability of class $j$ for example $i$. Similarly for class centroids $P^c = (p_{ji}^c) \in \real^{N \times N}$, support examples $P^s = (p_{ji}^s) \in \real^{N \times L}$ and queries $P^q = (p_{ji}^q) \in \real^{N \times M}$.


\subsection{Loss function: Class balancing or not}
\label{sec:mutual}

The set of all learnable parameters is $\Phi = \{C, G, B\}$ is optimized jointly using a mutual information loss~\cite{TIM, alphaTIM}. We distinguish between class-balanced and imbalanced tasks.

\subsubsection{Class-balanced tasks}

Following \cite{TIM}, we optimize parameters $\Phi$ using three loss terms. The first is standard average cross-entropy over the labeled support examples:
\begin{equation}
	L_\ce(P^s) = -\frac{1}{L} \sum_{i=1}^L \sum_{j=1}^N y^s_{ji} \log(p^s_{ji}).
\label{eq:ce}
\end{equation}
The second is the average, over queries, entropy of predicted class probability distributions per query
\begin{equation}
    \wb{\cH}(P^q) = -\frac{1}{M} \sum_{i=1}^M \sum_{j=1}^N p^q_{ji} \log(p^q_{ji}).
\label{eq:cond}
\end{equation}
This term aims at minimizing the uncertainty of the predicted probability distribution of every query, hence encouraging confident predictions. The third term is
\begin{equation}
    - \cH(\bar{\vp}^q) = \sum_{j=1}^N \bar{p}^q_j \log(\bar{p}^q_j),
\label{eq:entropy}
\end{equation}
where $\bar{p}^q_j = \frac{1}{M} \sum_{i=1}^M p^q_{ji}$ and $\bar{\vp}^q = (\bar{p}^q_j)_{j=1}^N =  P^q \vone_M \in \real^N$ is a vector representing the average predicted probability distribution of set $Q$. By maximizing its entropy, this term aims at maximizing its uncertainty, encouraging it to be uniform, hence balancing over classes.

The complete loss function to be minimized \wrt $\Phi$ is
\begin{equation}
    L_\bal = \lambda_3 L_\ce(P^s) + \lambda_2 \wb{\cH}(P^q) - \lambda_1 \cH(\bar{\vp}^q),
\label{eq:balanced}
\end{equation}
where $\lambda_1, \lambda_2, \lambda_3$ are scalar hyperparameters.

\subsubsection{Imbalanced tasks}

By encouraging the average predicted probability distribution to be uniform, the third term~\eq{entropy} is strongly biased towards class-balanced tasks. To make the loss more tolerant to imbalanced distributions, a relaxed version has been proposed based on the $\alpha$-divergence~\cite{alphaTIM}. In particular, the second~\eq{cond} and third term~\eq{entropy} become respectively
\begin{align}
	\wb{\cH}_\alpha(P^q) &= -\frac{1}{\alpha-1}
		\frac{1}{M} \sum_{i=1}^M \sum_{j=1}^N (p^q_{ji})^{\alpha} \label{eq:alpha-cond} \\
	-\cH_\alpha(\bar{\vp}^q) &= \frac{1}{\alpha-1}
		\sum_{j=1}^N (\bar{p}^q_j)^\alpha \label{eq:alpha-entropy}
\end{align}

In this case, the complete loss function~\eq{balanced} to be minimized with respect to $\Phi$ is modified as
\begin{equation}
    L_\imbal = \lambda_3 L_\ce(P^s) + \lambda_2 \wb{\cH}_\alpha(P^q) - \lambda_1 \cH_\alpha(\bar{\vp}^q).
\label{eq:imbalanced}
\end{equation}


\subsection{Manifold parameter optimization}
\label{sec:optimization}

In contrast to \cite{TIM} and \cite{alphaTIM}, rather than only optimizing the class centroids, we optimize the entire set of manifold parameters $\Phi$, which includes the class centroids $C$ as well as graph-specific parameters $G$~\eq{affinity} and $B$~\eq{adj_B}. We update $\Phi$ by minimizing~\eq{balanced} or~\eq{imbalanced} through any gradient-based optimization algorithm with learning rate $\epsilon$. The entire procedure from graph construction in \autoref{sec:graph} to manifold parameter optimization in \autoref{sec:optimization} is iterated for $r$ steps. Algorithm \ref{alg:main} summarizes the complete optimization procedure of our method.

\begin{algorithm}
\footnotesize

\DontPrintSemicolon
\SetFuncSty{textsc}
\SetDataSty{emph}
\definecolor{darkmidnightblue}{rgb}{0.0, 0.2, 0.4}
\definecolor{armygreen}{rgb}{0.29, 0.33, 0.13}
\definecolor{carmine}{rgb}{0.59, 0.0, 0.09}
\definecolor{chocolate}{rgb}{0.82, 0.41, 0.12}
\definecolor{pansypurple}{rgb}{0.47, 0.09, 0.29}
\newcommand{\commentsty}[1]{{\color{pansypurple}#1}}

\SetCommentSty{commentsty}
\SetKwComment{Comment}{$\triangleright$ }{}

\SetKwInOut{Input}{input}
\SetKwInOut{Output}{output}
\SetKwFunction{Graph}{graph}
\SetKwFunction{Affinity}{affinity}
\SetKwFunction{Adjacency}{adjacency}
\SetKwFunction{Features}{features}
\SetKwFunction{Centroids}{centroids}
\SetKwFunction{Initialize}{initialize}

\SetKwFunction{Label}{label}
\SetKwFunction{LP}{lp}
\SetKwFunction{Power}{power}
\SetKwFunction{Balance}{balance}
\SetKwFunction{Sinkhorn}{Sinkhorn}
\SetKwFunction{Predict}{predict}
\SetKwFunction{Clean}{clean}
\SetKwFunction{Augment}{augment}
\SetKwFunction{Softmax}{softmax}
\SetKwFunction{Loss}{loss}
\SetKwFunction{Set}{set}
\SetKwFunction{Update}{update}

\Input{ Pre-trained backbone $f_{\theta}$}
\Input{ labeled support set $S$ with $\card{S}=L$}
\Input{ unlabeled query set $Q$ with $\card{Q}=M$}
\BlankLine

$(V^s, V^q) \gets (f_{\theta}(S), f_{\theta}(Q))$\\
$C \gets \Centroids (V^s)$ \Comment*{class centroids~\eq{centroid}}
$\cV \gets \{C, V^s, V^q\}$\\
$(G, B) \gets \Initialize()$\\
$\Phi \gets \{C, G, B\}$\\

\For {$r$ steps}
{
	$A \gets \Affinity(\cV; G,k)$ \Comment*{affinity matrix~\eq{affinity}}
 	$W \gets \frac{1}{2}(A+A^T)$ \Comment*{symmetric adjacency matrix}
 	$W_B \gets W \circ B$ \Comment*{scaled adjacency matrix~\eq{adj_B}}
 	$\cW \gets D^{-1/2} W_B D^{-1/2}$ \Comment*{adjacency matrix~\eq{adj}}
	$Y \gets (I_N\ \ \vzero_{N \times L}\ \ \vzero_{N \times M})$ \Comment*{label matrix~\eq{label}}
	$Z \gets Y (I - \beta \cW)^{-1}$ \Comment*{label propagation~\eq{lp}}
	$P \gets \Softmax(Z)$ \Comment*{class probabilities~\eq{softmax}}
	$L_{\mathrm{bal}}/L_{\mathrm{imbal}} \gets \Loss(P; \Phi)$ \Comment*{loss function~\eq{balanced} or \eq{imbalanced}}
	$\Phi \gets \Update(\Phi; L_{\mathrm{bal}}/L_{\mathrm{imbal}})$
}
\textbf{return} $P^q$
\caption{Adaptive Manifold (\ours).}
\label{alg:main}
\end{algorithm}


\subsection{Transductive Inference}
\label{sec:transductive_inference}

Upon convergence of the optimization of manifold parameters $\Phi$, we obtain the final query probability matrix $P^q$~\eq{prob} and for each query $x_i^q \in Q$, we predict the \emph{pseudo-label}
\begin{equation}
	\hat{y}_i^q = \arg\max_j p_{ji}^q
\label{eq:argmax}
\end{equation}
corresponding to the maximum element of the $i$-th column of matrix $P^q$.

\section{Experiments}
\label{sec:experiments}

\newcommand{\ci}[1]{{\tiny $\pm$#1}}
\newcommand{\cip}{\phantom{\ci{0.00}}}
\newcommand{\cim}{\ci{\alert{0.00}}}
\input{fig/data/sample}

\subsection{Setup}
\label{sec:setup}

\paragraph{Datasets}

We experiment on three commonly used few-shot learning benchmark datasets, namely \emph{mini}ImageNet~\cite{matchingNets}, \emph{tiered}ImageNet~\cite{sslmeta} and CUB~\cite{closerlook}. In state of the art comparisons in the balanced setting, we also use CIFAR-FS~\cite{closerlook, cifar100db}.

\paragraph{Backbones}

We use the three pre-trained backbones from the publicly available code~\cite{alphaTIM}, namely ResNet-18, WideResNet-28-10 (WRN-28-10) and DenseNet-121. All are trained using standard cross entropy loss on $D_\base$ for 90 epochs with learning rate 0.1, divided by 10 at epochs 45 and 66. Color jittering, random cropping and random horizontal flipping augmentations are used at training. We also carry out experiments using the publicly available code and pre-trained WRN-28-10 backbones provided by~\cite{ilpc}.

\paragraph{Tasks}

Unless otherwise stated, we consider $N$-way, $K$-shot tasks with $N = 5$ randomly sampled classes from $C_\novel$ and $K \in \{1, 5\}$ random labeled examples for the support set $S$. The query set $Q$ contains $M = 75$ query examples in total. In the balanced setting, there are $\frac{M}{N}=\frac{75}{5} = 15$ queries per class. In the imbalanced setting, the total number of queries remains $M=75$. Following~\cite{alphaTIM}, we sample imbalanced tasks by modeling the proportion of examples from each class in $Q$ as a vector $\vpi = (\pi_1, \dots, \pi_N)$ sampled from a symmetric Dirichlet distribution $\Dir(\gamma)$ with parameter $\gamma = 2$. We follow \cite{alphaTIM} and \cite{ilpc}, performing 10000 and 1000 tasks respectively when using the code and settings of each work.

\paragraph{Implementation details}

Our implementation is in Pytorch~\cite{pytorch}. We carry out experiments for balanced and imbalanced transductive few-shot learning using the publicly available code provided by~\cite{alphaTIM}\footnote{\url{https://github.com/oveilleux/Realistic_Transductive_Few_Shot}}. For additional experiments in the balanced setting, we use the publicly available code provided from~\cite{ilpc}\footnote{\url{https://github.com/MichalisLazarou/iLPC}}. We used Adam optimizer in for the manifold parameter optimization \autoref{sec:optimization}.

\paragraph{Hyperparameters}

Following~\cite{alphaTIM}, we keep the same values of hyper-parameters $\tau$, $\lambda_1$, $\lambda_2$, $\lambda_3$, $\epsilon$ and $r$. We set $\epsilon=0.0001$, $r = 1000$, $\tau = 15$~\eq{softmax}. In the imbalanced setting we set $\lambda_1= \lambda_2=\lambda_3=1$, while in the balanced setting we set $\lambda_1=\lambda_3=1$ and $\lambda_2 = 10$. Regarding hyperparameter $\alpha$~\eq{alpha-cond},\eq{alpha-entropy} we ablate it in section \autoref{sec:ablation} and set $\alpha = 2$ for 1-shot and $\alpha = 5$ for 5-shot for all experiments unless stated otherwise. For label propagation, we set $k = 20$~\eq{edges} for 1-shot and $k = 10$ for 5-shot; we initialize $G = J_T$~\eq{affinity} and $B = J_T$~\eq{adj_B} where $J_T$ is the $T \times T$ all-ones matrix; we initialize $\beta = 0.8$~\eq{lp} for 1-shot and $\beta = 0.9$ for 5-shot. We optimized $k$, $\beta$ and the initialization of $G$ and $B$ on the \emph{mini}ImageNet validation set. To avoid hyperparameter overfitting, \emph{all hyper-parameters are kept fixed across all datasets and backbones}.

\paragraph{Baselines}

In the imbalanced setting, we compare our method against the state of the art method $\alpha$-TIM~\cite{alphaTIM}, basing our experiments on the publicly available code and comparing against all methods implemented in that code. In the balanced setting, we reproduce results of all methods provided in the official code from~\cite{alphaTIM} and compare our method against them. Furthermore regarding the balanced setting, we compare our method against other state of the art methods such as \cite{ilpc, ease} by using the official code from~\cite{ilpc}.


\paragraph{Feature pre-processing}

We experiment with two commonly used feature pre-processing methods, denoted as $\eta$ in \autoref{sec:problem}, namely $\ell_2$-normalization and the method used in \cite{pt_map, ilpc}, which we refer to as \plc. $\ell_2$-normalization is defined as $\frac{\vv}{||\vv||_2}$ for $\vv \in V$. \plc, standing for \emph{power transform, $\ell_2$-normalization, centering}, performs elementwise power transform $\vv^{\frac{1}{2}}$ for $\vv \in V$, followed by $\ell_2$-normalization and centering, subtracting the mean over $V$.

In the balanced and imbalanced settings respectively, we refer to our method as \ours, $\alpha$-\ours when using $\ell_2$-normalization and as \oursplc, $\alpha$-\oursplc when using \plc pre-processing. TIM~\cite{TIM} and $\alpha$-TIM~\cite{alphaTIM} use only $\ell_2$-normalization originally. For fair comparisons, we also use \plc pre-processing on TIM and $\alpha$-TIM, referring to them as TIM$_\textsc{plc}$, $\alpha$-TIM$_{\textsc{plc}}$ in the balanced and imbalanced settings respectively.

\begin{table*}
\small
\centering
\begin{tabular}{ccccccccccccccc}
\toprule
& & & & & \mc{4}{\Th{Imbalanced}}   & \mc{4}{\Th{Balanced}}              \\
\mc{5}{\Th{Components}} & \mc{2}{\Th{ResNet-18}}                   & \mc{2}{\Th{WRN-28-10}}      & \mc{2}{\Th{ResNet-18}}                   & \mc{2}{\Th{WRN-28-10}}              \\
$\NN_k$ & $C$ & $G$ & $B$ & \plc &  1-shot              & 5-shot              & 1-shot              & 5-shot       & 1-shot              & 5-shot              & 1-shot              & 5-shot                    \\ \midrule
    &      &       &  &   & 60.21\ci{0.27}       &   74.24\ci{0.21}   &  63.34\ci{0.27}    & 76.19\ci{0.21}       & 59.09\ci{0.21}      & 71.54\ci{0.19}     & 62.38\ci{0.21}     & 73.46\ci{0.19} \\
\ch &      &      &  &   & 63.95\ci{0.27}     & 81.15\ci{0.17}      & 67.14\ci{0.27}      &  83.40\ci{0.16}    & 63.82\ci{0.22}      & 80.47\ci{0.15}     & 67.22\ci{0.21}      & 82.58\ci{0.16}  \\
\ch & \ch  &     &  &           & 68.57\ci{0.28}      &  82.69\ci{0.16}    & 71.22\ci{0.26}    & 84.74\ci{0.16}  & 73.43\ci{0.23}      &   84.37\ci{0.14}       & 75.94\ci{0.22}     & 86.55\ci{0.13}   \\
\ch &   \ch   & \ch     &  &        & 70.16\ci{0.29}      & 82.62\ci{0.17}    &    72.89\ci{0.28} & 84.89\ci{0.16}  &  75.59\ci{0.27}      &  84.80\ci{0.15}       &  78.72\ci{0.25}     & 87.11\ci{0.13}   \\
\ch & \ch  &     &  \ch  &          & 69.11\ci{0.29}      & 82.97\ci{0.16}    & 71.64\ci{0.28}    & 85.16\ci{0.15}  & 74.85\ci{0.25}      &  84.66\ci{0.14}       & 77.70\ci{0.23}     &  86.91\ci{0.13}   \\
  \ch  & \ch  & \ch    & \ch & &  \tb{70.24}\ci{0.29}      & 82.71\ci{0.17}    &   \tb{73.22}\ci{0.29}   &  85.00\ci{0.16}  & 76.06\ci{0.28}      &  84.82\ci{0.15}       & 79.37\ci{0.26 }    & 87.12\ci{0.13}   \\
    \ch  & \ch  & \ch    & \ch & \ch & 69.97\ci{0.29}      & \tb{83.31}\ci{0.17}    & 71.98\ci{0.29}    & \tb{85.66}\ci{0.15}  & \tb{77.35}\ci{0.27}      &  \tb{85.47}\ci{0.14}       &  \tb{80.99}\ci{0.26}     &  \tb{87.86}\ci{0.13}   \\

\bottomrule
\end{tabular}
\vspace{6pt}
\caption{\emph{Ablation study of algorithmic components} of both balanced and imbalanced versions of our method \ours on \emph{mini}ImageNet. $\NN_k$: $k$-nearest neighbour graph; otherwise, complete graph. $C$: learnable class centroids. $G$: learnable pairwise scaling factors $G$~\eq{affinity}. $B$: learnable adjacency matrix $B$~\eq{adj_B}. \plc: feature pre-processing as defined in \autoref{sec:setup}.}
\label{tab:ablation}
\end{table*}

\paragraph{Reporting results}

In every table we denote the best performing results with bold regardless the pre-processing method used. Nevertheless, since our work is influenced by~\cite{TIM} and~\cite{alphaTIM}, we also compare with these two methods under the same feature pre-processing settings. In \autoref{tab:soa-imbalanced}, \autoref{tab:imbalanced_cub}, \autoref{tab:soa-balanced}, \autoref{tab:soa-balanced_cub} and \autoref{tab:densenet}, we use the code by~\cite{alphaTIM}, reporting the mean accuracy over 10000 tasks \cite{alphaTIM}. In \autoref{tab:trans_balanced_ilpc}, we use the code by~\cite{ilpc}, reporting the mean accuracy and 95$\%$ confidence interval over 1000 tasks. In the ablation study in \autoref{tab:ablation} we use the code by~\cite{alphaTIM}, however, since we ablate our own method, we report both the mean accuracy and the $95\%$ confidence interval.

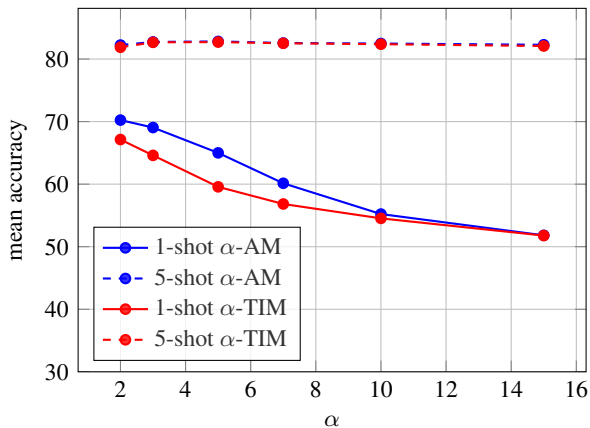
\begin{figure}
\extfig{pca}
{
\begin{tikzpicture}
\begin{axis}[
	width=\linewidth,
	height=0.77\linewidth,
	font=\small,
	ymin=30,
	xlabel={$\alpha$},
	ylabel={mean accuracy},
	legend pos={south west},
]
\addplot[blue, mark=*] table[x=alphavalues, y=oneshot] \alphagraph; \leg{1-shot $\alpha$-\ours}
\addplot[blue, dashed, mark=*]  table[x=alphavalues,y=fiveshot]  \alphagraph; \leg{5-shot $\alpha$-\ours}
\addplot[red, mark=*] table[x=alphavalues, y=timone] \alphagraph; \leg{1-shot $\alpha$-TIM}
\addplot[red, dashed, mark=*]  table[x=alphavalues,y=timfive]  \alphagraph; \leg{5-shot $\alpha$-TIM}
\end{axis}
\end{tikzpicture}
}
\caption{\emph{Effect of parameter $\alpha$} on $\alpha$-\ours and $\alpha$-TIM, \emph{mini}ImageNet 1-shot and 5-shot.}
\label{fig:alpha_fig}
\end{figure}

\subsection{Ablation study}
\label{sec:ablation}

\paragraph{Ablation of hyper-parameter $\alpha$}

\autoref{fig:alpha_fig} ablates $\alpha$-\ours and $\alpha$-TIM with respect to $\alpha$. It can be seen that the value of $\alpha$ has a lot more effect in the 1-shot than in the 5-shot setting. Also, $\alpha$ behaves similarly for both $\alpha$-\ours and $\alpha$-TIM. Nevertheless, in the majority of the cases $\alpha$-\ours outperforms $\alpha$-TIM. The optimal value of $\alpha$ is $2$ for 1-shot and $5$ for 5-shot. Therefore we set $\alpha = 2$ and $\alpha = 5$ for all 1-shot and 5-shot imbalanced experiments respectively unless stated otherwise.

\paragraph{Algorithmic components}

We ablate all components of our method under both the imbalanced~\eq{imbalanced} and balanced~\eq{balanced} settings in \autoref{tab:ablation}. As it can be seen from the first and second rows, using a $k$-nearest neighbour graph gives significant improvement over using a dense graph. Adapting the centroids, $C$, brings further substantial improvement. Adapting the centroids, $C$, along with either $G$ or $B$ brings further performance improvement. Adapting both manifold parameters $G$ and $B$ along with $C$ provides better performance than just adapting either $G$ or $B$ in most experiments, especially in the balanced setting. Using \plc pre-processing yields further performance improvement except in the 1-shot imbalanced setting.

\begin{table}
\small
\centering
\setlength\tabcolsep{5pt}
\begin{tabular}{lcccc}
\toprule
\mr{2}{\Th{Method}}                                        & \mc{2}{\Th{\emph{mini}ImageNet}}  & \mc{2}{\Th{\emph{tiered}ImageNet}}     \\
                                               & 1-shot              & 5-shot & 1-shot              & 5-shot   \\ \midrule
                                       \mc{5}{\Th{ResNet-18}}  \\ \midrule

Entropy-min~\cite{entropy-min}              & 58.50     & 74.80    & 61.20     & 75.50\\
LR+ICI~\cite{lrici}                             & 58.70     & 73.50    &  74.60    &   85.10\\
PT-MAP~\cite{pt_map}                           & 60.10     & 67.10   & 64.10     & 70.00 \\
LaplacianShot~\cite{laplacianshot}             & 65.40     & 81.60   & 72.30     & 85.70 \\
BD-CSPN~\cite{BDCSPN}                 & 67.00     & 80.20   & 74.10     & 84.80 \\
TIM~\cite{TIM}                                & 67.30     & 79.80   & 74.10     & 84.10 \\ \midrule
$\alpha$-TIM~\cite{alphaTIM}                   & 67.40     & 82.50    & 74.40    & 86.60\\
$\alpha$-TIM$_{\textsc{plc}}$*~\cite{alphaTIM}    & 63.38    &   82.80   & 70.17     & 86.82\\ 
\rowcolor{LightCyan} $\alpha$-\ours                       & \tb{70.24}     & 82.71  & \tb{77.28}    & 86.97 \\
\rowcolor{LightCyan} $\alpha$-\oursplc                    &  69.97     & \tb{83.31}  & 76.44   & \tb{87.19} \\
\midrule
                                       \mc{5}{\Th{WRN-28-10}}  \\ \midrule
Entropy-min \cite{entropy-min}         & 60.40    & 76.20 & 62.90     & 77.30      \\
PT-MAP~\cite{pt_map}                         & 60.60     & 66.80   & 65.10     & 71.00  \\
LaplacianShot~\cite{laplacianshot}             & 68.10     & 83.20   & 73.50     & 86.80 \\
BD-CSPN~\cite{BDCSPN}                   & 70.40     & 82.30   &  75.40    & 85.90 \\
TIM~\cite{TIM}                                 & 69.80     & 81.60    & 75.80    & 85.40 \\ \midrule
$\alpha$-TIM~\cite{alphaTIM}                   & 69.80     & 84.80     & 76.00   & 87.80 \\
$\alpha$-TIM$_{\textsc{plc}}$*~\cite{alphaTIM}                  & 66.50     & 85.12    & 71.97     &  88.28\\ 
\rowcolor{LightCyan} $\alpha$-\ours                              & \tb{73.22}     & 85.00  & \tb{78.94}     & 88.44\\

\rowcolor{LightCyan} $\alpha$-\oursplc                             & 71.98     & \tb{85.66}   & 78.75    & \tb{88.69} \\
\bottomrule
\end{tabular}
\vspace{6pt}
\caption{\emph{Imbalanced transductive inference} on \emph{mini}ImageNet and \emph{tiered}ImageNet. Results as reported by~\cite{alphaTIM}. *: Results were reproduced using the official code provided by~\cite{alphaTIM}.
}
\label{tab:soa-imbalanced}
\end{table}

\begin{table}
\small
\centering
\setlength\tabcolsep{5pt}
\begin{tabular}{lcc}
\toprule
\mr{2}{\Th{Method}} & \mc{2}{\Th{CUB}}       \\
           & 1-shot              & 5-shot         \\ \midrule
\mc{3}{\Th{ResNet-18}} \\ \midrule
PT-MAP~\cite{pt_map}                    & 65.10     & 71.30      \\
Entropy-min~\cite{entropy-min}           & 67.50     & 82.90          \\
LaplacianShot~\cite{laplacianshot}       & 73.70     & 87.70           \\
BD-CSPN~\cite{BDCSPN}                   & 74.50     & 87.10           \\
TIM~\cite{TIM}                            & 74.80     & 86.90        \\ \midrule
$\alpha$-TIM~\cite{alphaTIM}             & 75.70     & 89.80       \\
$\alpha$-TIM$_{\textsc{plc}}$*~\cite{alphaTIM}        & 70.95   & 89.56   \\
\rowcolor{LightCyan} $\alpha$-\ours             & \tb{79.92} & 89.83 \\
\rowcolor{LightCyan} $\alpha$-\oursplc          & 78.62 & \tb{89.86} \\

\bottomrule

\end{tabular}
\vspace{6pt}
\caption{\emph{Imbalanced transductive inference} on CUB. Results as reported by~\cite{alphaTIM}. *: Results were reproduced using the official code provided by~\cite{alphaTIM}.}
\label{tab:imbalanced_cub}
\end{table}

\begin{table*}
\small
\centering
\setlength\tabcolsep{5pt}
\begin{tabular}{lcccccccc}
\toprule
\mr{2}{\Th{Method}}                             & \mc{2}{\Th{\emph{mini}ImageNet}}          & \mc{2}{\Th{\emph{tiered}ImageNet}}        & \mc{2}{\Th{Cifar-FS}}                     & \mc{2}{\Th{CUB}}                          \\
                                               & 1-shot              & 5-shot              & 1-shot              & 5-shot              & 1-shot              & 5-shot              & 1-shot              & 5-shot              \\ \midrule
                                        \mc{9}{\Th{WRN-28-10}} \\
\midrule
PT+MAP~\cite{pt_map}$^*$                    & 82.88\ci{0.73}      & 88.78\ci{0.40}      & 88.15\ci{0.71}      & 92.32\ci{0.40}      & 86.91\ci{0.72}      & 90.50\ci{0.49}      & 91.37\ci{0.61}      & 93.93\ci{0.32}      \\
\ilpc~\cite{ilpc}$^*$          & 83.05\ci{0.79} & 88.82\ci{0.42} & 88.50\ci{0.75} & 92.46\ci{0.42} & 86.51\ci{0.75} & 90.60\ci{0.48} & 91.03\ci{0.63} & 94.11\ci{0.30}\\
EASE+SIAMESE~\cite{ease}$^\dagger$          & \tb{83.44}\ci{0.77} & 88.66\ci{0.43} & \tb{88.69}\ci{0.73} & 92.47\ci{0.41} & 86.71\ci{0.77} & 90.28\ci{0.51} & 91.44\ci{0.63} & 93.85\ci{0.32} \\
EASE+SIAMESE$_\textsc{plc}$~\cite{ease}$^\dagger$          & 82.13\ci{0.81} & 87.34\ci{0.46}& 88.42\ci{0.73} & 92.19\ci{0.41} & 86.74\ci{0.78} & 90.22\ci{0.51} & \tb{91.49}\ci{0.63} & 93.32\ci{0.32} \\
TIM \cite{TIM}        & 77.65\ci{0.72} & 88.21\ci{0.40}  & 83.88\ci{0.74} & 91.89\ci{0.41} & 82.63\ci{0.70} & 90.28\ci{0.46} & 87.50\ci{0.62} & 93.59\ci{0.30}\\
TIM$_{\textsc{plc}}$ \cite{TIM}    & 75.77\ci{0.67} & 88.37\ci{0.40}  & 83.22\ci{0.70} & 92.13\ci{0.40} & 80.52\ci{0.70} & 90.25\ci{0.46} & 85.58\ci{0.61} & 93.48\ci{0.31}\\
\rowcolor{LightCyan} \ours            & 80.74\ci{0.81} & 87.75\ci{0.42}  & 86.38\ci{0.78} &  91.85\ci{0.85} & 85.93\ci{0.74} & 90.13\ci{0.47} & 90.24\ci{0.65} & 93.43\ci{0.30} \\
\rowcolor{LightCyan} \oursplc        & 83.40\ci{0.74} & \tb{89.08}\ci{0.40}  &  88.31\ci{0.73} &   \tb{92.60}\ci{0.39} & \tb{86.91}\ci{0.74} & \tb{90.80}\ci{0.46} & 91.32\ci{0.60} & \tb{94.14}\ci{0.29} \\\bottomrule
\end{tabular}
\vspace{6pt}
\caption{\emph{Balanced transductive inference state of the art}. Results were reproduced using the official code provided by~\cite{ilpc}. *: Results as reported by~\cite{ilpc}. $\dagger$: Our reproduction with official code from~\cite{ease}.}
\label{tab:trans_balanced_ilpc}
\end{table*}
\subsection{Comparison of state of the art}

\paragraph{Imbalanced transductive few-shot learning}

\autoref{tab:soa-imbalanced} and \autoref{tab:imbalanced_cub} show that our method achieves new state of the art performance using both ResNet-18 and WRN-28-10 on all three datasets and both 1-shot and 5-shot settings. Impressively, we improve the 1-shot state of the art in all cases significantly, by as much as $4.2\%$ on CUB with ResNet-18. Even though we outperform $\alpha$-TIM without \plc pre-processing in every experiment, \plc brings further improvement in 5-shot, while not being beneficial in 1-shot. Interestingly, \plc pre-processing does not have the same effect on $\alpha$-TIM, providing only marginal improvement in the 5-shot while being detrimental in the 1-shot.
\begin{table}
\small
\centering
\setlength\tabcolsep{5pt}
\begin{tabular}{lcccc}
\toprule
 \mr{2}{\Th{Method}}                                      & \mc{2}{\Th{\emph{mini}ImageNet}}  & \mc{2}{\Th{\emph{tiered}ImageNet}}        \\

                                               & 1-shot              & 5-shot & 1-shot              & 5-shot           \\ \midrule
                                       \mc{5}{\Th{ResNet-18}}  \\ \midrule

PT-MAP~\cite{pt_map}                           & 76.88    & 85.18   & 82.89    & 88.64          \\
LaplacianShot~\cite{laplacianshot}             & 70.24    & 82.10   & 77.28    & 86.22          \\
BD-CSPN~\cite{BDCSPN}                           & 69.36    & 82.06   & 76.36    & 86.18         \\
TIM~\cite{TIM}                                & 73.81    & 84.91   & 80.13    & 88.61       \\
TIM$_{\textsc{plc}}$~\cite{TIM}                     & 69.33    & 84.53  & 76.36   & 88.33 \\
\rowcolor{LightCyan} \ours                     &  76.06      &  84.82 &82.42 & 88.61     \\
\rowcolor{LightCyan} \oursplc                   & \tb{77.35} &  \tb{85.47} & \tb{83.40} &\tb{89.07}     \\
\midrule
                                       \mc{5}{\Th{WRN-28-10}}  \\ \midrule
PT-MAP~\cite{pt_map}                         &  80.35   & 87.37   & 84.84    & 89.86         \\
LaplacianShot~\cite{laplacianshot}             &  72.91    & 83.85   & 78.85    & 87.27         \\
BD-CSPN~\cite{BDCSPN}                   & 72.16   & 83.78   & 77.88   & 87.23       \\
TIM~\cite{TIM}                       &           77.78    & 87.43   & 82.28    & 89.84          \\
TIM$_{\textsc{plc}}$~\cite{TIM}                     & 73.52    & 86.95   & 78.23    & 89.56 \\
\rowcolor{LightCyan} \ours                              & 79.37  & 87.12 & 84.07 &  89.69     \\
\rowcolor{LightCyan} \oursplc                           & \tb{80.99}     &  \tb{87.86} & \tb{85.26} & \tb{90.30}     \\

\bottomrule
\end{tabular}
\vspace{6pt}
\caption{\emph{Balanced transductive inference} on \emph{mini}ImageNet and \emph{tiered}ImageNet. All results were reproduced using the official code provided by~\cite{alphaTIM}.
}
\label{tab:soa-balanced}
\end{table}
\begin{table}
\small
\centering
\setlength\tabcolsep{5pt}
\begin{tabular}{lcc}
\toprule
\mr{2}{\Th{Method}}                            & \mc{2}{\Th{CUB}}        \\
     & 1-shot              & 5-shot            \\ \midrule
PT-MAP~\cite{pt_map}                    & 86.05    & 91.28        \\
LaplacianShot~\cite{laplacianshot}      & 79.55    & 88.96         \\
BD-CSPN~\cite{BDCSPN}                   & 78.52   & 89.02        \\
TIM~\cite{TIM}                          & 82.87     & 91.58       \\
TIM$_{\textsc{plc}}$~\cite{TIM}                     & 77.69    & 91.17 \\
\rowcolor{LightCyan} \ours                                   & 85.59        & 91.24\\
\rowcolor{LightCyan} \oursplc                                & \tb{86.64}         & \tb{91.78} \\
\bottomrule
\end{tabular}
\vspace{6pt}
\caption{\emph{Balanced transductive inference} on CUB. All results were reproduced using the official code provided by~\cite{alphaTIM}.}
\label{tab:soa-balanced_cub}
\end{table}

\paragraph{Balanced transductive few-shot learning}

Tables~\ref{tab:soa-balanced} and~\ref{tab:soa-balanced_cub} show that \oursplc outperforms all other methods, with its closest competitor being PT-MAP~\cite{pt_map}. Notably, our superiority is not due to pre-processing because PT-MAP also uses \plc. \oursplc also significantly outperforms both versions of TIM. Interestingly, the performance of TIM always drops when \plc pre-processing is used, while \ours always improves. Even without \plc, \ours significantly outperforms TIM by $2-4\%$ in 1-shot, while being on par or slightly worse by $0.1-0.3\%$ in 5-shot.

Since the official code provided by~\cite{alphaTIM} does not provide more recent methods in the balanced setting, such as~\cite{ilpc} and~\cite{ease}, we use the publicly available code and pre-trained WRN-28-10 provided by~\cite{ilpc} to compare \ours with the state of the art methods: TIM, EASE+SIAMESE~\cite{ease}, PT+MAP~\cite{pt_map} and iLPC~\cite{ilpc}. \autoref{tab:trans_balanced_ilpc} shows that \oursplc outperforms all methods in the majority of the experiments. Again, our superiority is not due to pre-processing since we provided results for TIM and EASE+SIAMESE using \plc while PT-MAP and iLPC use \plc as part of their method.

\subsection{Effect of unlabeled data}

\begin{figure}
\extfig{pca}
{
\begin{tikzpicture}
\begin{axis}[
	width=\linewidth,
	height=0.62\linewidth,
	font=\small,
	ymin=82,
	xlabel={number of unlabeled queries},
	ylabel={mean accuracy},
	legend pos={south east},
]
\addplot[blue, mark=*] table[x=unlabeled, y=ours5] \unlabeledone; \leg{$\alpha$-\ours}
\addplot[red, mark=*]  table[x=unlabeled,y=tim5]  \unlabeledone; \leg{$\alpha$-TIM}
\end{axis}
\end{tikzpicture}
}
\caption{\emph{Effect of number of unlabeled queries $M$} on $\alpha$-\ours and $\alpha$-TIM, \emph{mini}ImageNet 5-shot.}
\label{fig:num-unlabeled5}
\end{figure}
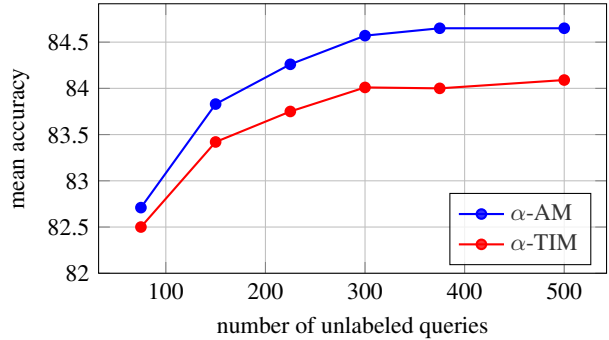

We investigate the effect of the quantity of unlabeled queries $M$, comparing $\alpha$-\ours against $\alpha$-TIM. We do not use \plc pre-processing here, \emph{which is beneficial to $\alpha$-TIM}. \autoref{fig:num-unlabeled} shows that in the 1-shot setting, $\alpha$-\ours outperforms $\alpha$-TIM significantly by as much as $3.7\%$ when $M = 300$ and generally the performance gap increases as the number of unlabeled data increases. \autoref{fig:num-unlabeled5} shows that also in the 5-shot setting, as the number of unlabeled queries increases, our algorithm outperforms $\alpha$-TIM with an increasing performance gap. This can be attributed to the fact that $\alpha$-\ours leverages the data manifold through the $k$-nearest neighbour graph while $\alpha$-TIM works in Euclidean space.

\begin{figure}
\extfig{pca}
{
\begin{tikzpicture}
\begin{axis}[
	width=\linewidth,
	height=0.62\linewidth,
	font=\small,
	ymin=67,
	xlabel={number of unlabeled queries},
	ylabel={mean accuracy},
	legend pos={south east},
]
\addplot[blue, mark=*] table[x=unlabeled, y=ours] \unlabeledone; \leg{$\alpha$-\ours}
\addplot[red, mark=*]  table[x=unlabeled,y=tim]  \unlabeledone; \leg{$\alpha$-TIM}
\end{axis}
\end{tikzpicture}
}
\caption{\emph{Effect of number of unlabeled queries $M$} on $\alpha$-\ours and $\alpha$-TIM, \emph{mini}ImageNet 1-shot.}
\label{fig:num-unlabeled}
\end{figure}
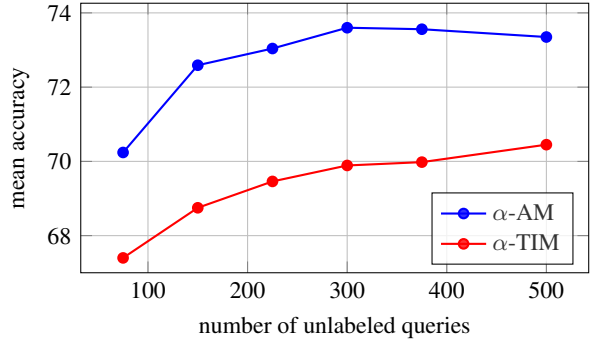

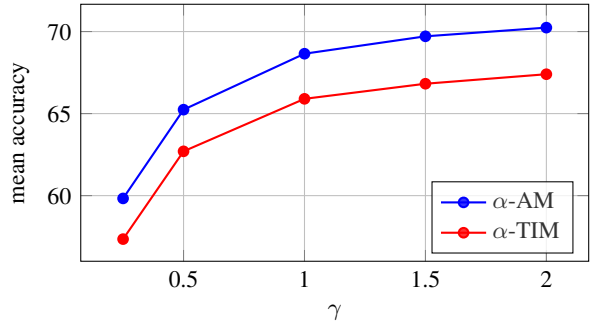
\begin{figure}
\extfig{pca}
{
\begin{tikzpicture}
\begin{axis}[
	width=\linewidth,
	height=0.6\linewidth,
	font=\small,
	ymin=56,
	xlabel={$\gamma$},
	ylabel={mean accuracy},
	legend pos={south east},
]
\addplot[blue, mark=*] table[x=dirichlet, y=ours] \alphadirichlet; \leg{$\alpha$-\ours}
\addplot[red, mark=*]  table[x=dirichlet,y=tim]  \alphadirichlet; \leg{$\alpha$-TIM}
\end{axis}
\end{tikzpicture}
}
\caption{\emph{Effect of class imbalance parameter $\gamma$} in $\Dir(\gamma)$ on $\alpha$-\ours and $\alpha$-TIM, \emph{mini}ImageNet 1-shot. Class distributions are more imbalanced with lower $\gamma$.}
\label{fig:gamma_fig}
\end{figure}

\subsection{Robustness against imbalance}

We investigate the effect of increasing the class imbalance in $Q$ by decreasing the value of $\gamma$ used in $\Dir(\gamma)$. \autoref{fig:gamma_fig} shows that, while the performance of both $\alpha$-\ours and $\alpha$-TIM drops as the classes become more imbalanced, $\alpha$-\ours consistently outperforms $\alpha$-TIM.

\subsection{Other backbones}

\autoref{tab:densenet} shows that by using the DenseNet-121 backbone, \oursplc outperforms $\alpha$-TIM and $\alpha$-TIM$_\textsc{plc}$ in both 1-shot and 5-shot settings. As in the previous experiments, we observe a significant performance gap of roughly $3.5\%$ in the 1-shot setting.

\begin{table}
\small
\centering
\begin{tabular}{lcccc}
\toprule
\mr{2}{\Th{Method}}   & \mc{2}{\Th{\emph{mini}ImageNet}}          & \mc{2}{\Th{\emph{tiered}ImageNet}} \\
           & 1-shot              & 5-shot              & 1-shot              & 5-shot\\ \midrule
                        \mc{5}{\Th{DenseNet-121}} \\ \midrule
$\alpha$-TIM~\cite{alphaTIM}     & 70.41  & 85.58 &  76.55 & 88.33     \\
$\alpha$-TIM$_{\textsc{plc}}$~\cite{alphaTIM}     & 67.56  & 86.26 &  74.56 &  88.68   \\
\rowcolor{LightCyan}$\alpha$-\ours              & 73.67 & 85.47 &  79.95 &89.34 \\
\rowcolor{LightCyan}$\alpha$-\oursplc           & \tb{73.98}     & \tb{86.76} &   \tb{79.99} & \tb{89.73}\\ 
\bottomrule
\end{tabular}
\vspace{6pt}
\caption{\emph{Imbalanced transductive inference} on \emph{mini}ImageNet and \emph{tiered}ImageNet using the DenseNet-121 backbone. All results were reproduced using the official code provided by~\cite{alphaTIM}.}
\label{tab:densenet}
\end{table}

\section{Conclusion}
\label{sec:conclusion}

In this work we propose a novel method named as \emph{Adaptive Manifold}, \ours, that achieves new state of the art performance in imbalanced transductive few-shot learning, while outperforming several state of the art methods in the traditional balanced transductive few-shot learning. \ours combines the complementary strengths of K-means-like iterative class centroid updates and exploiting the underlying data manifold through label propagation. Leveraging manifold class similarities to measure class probabilities for the unlabeled query examples and optimizing the manifold parameters through the loss function proposed by~\cite{alphaTIM}, we achieve a new state of the art performance in the imbalanced setting on different datasets using multiple backbones, outperforming previous methods by a large margin, especially in the 1-shot setting.  The robustness of our method is validated by our findings that it can be combined effectively with \plc pre-processing and that it can outperform its competitors in other settings such as with more unlabeled query examples and as well as the standard balanced few-shot setting.

{\small
\bibliographystyle{ieee_fullname}
\bibliography{tex/references}
}

\end{document}